\documentclass[letterpaper, 10 pt, conference]{IEEEtran}  

\IEEEoverridecommandlockouts                              

\usepackage{cite}
\usepackage{textcomp}
\usepackage{xcolor}
\def\BibTeX{{\rm B\kern-.05em{\sc i\kern-.025em b}\kern-.08em
    T\kern-.1667em\lower.7ex\hbox{E}\kern-.125emX}}

\usepackage{graphicx} 
\usepackage{amsmath} 
\usepackage{amssymb}  
\usepackage[ruled]{algorithm2e} 
\SetAlFnt{\footnotesize}
\usepackage[hyphens]{url}
\usepackage{hyperref}  
\usepackage{comment}
\usepackage[caption=false]{subfig}
\usepackage{copyright}

\title{
Simulation-Based Counterfactual Causal Discovery on Real World Driver Behaviour 
}

\author{
\IEEEauthorblockN{1\textsuperscript{st} Rhys Howard}
\IEEEauthorblockA{\textit{Oxford Robotics Institute} \\ \textit{Dept. of Engineering Science} \\
\textit{University of Oxford}\\
rhyshoward@live.com}
\and
\IEEEauthorblockN{2\textsuperscript{nd} Lars Kunze}
\IEEEauthorblockA{\textit{Oxford Robotics Institute} \\ \textit{Dept. of Engineering Science} \\
\textit{University of Oxford}\\
lars@robots.ox.ac.uk}
\thanks{This work was supported by the EPSRC project RAILS (grant reference: EP/W011344/1) and the Oxford Robotics Institute project RobotCycle.}
}

\begin{document}

\maketitle
\thispagestyle{empty}
\pagestyle{empty}

\begin{abstract}

Being able to reason about how one's behaviour can affect the behaviour of others is a core skill required of intelligent driving agents. Despite this, the state of the art struggles to meet the need of agents to discover causal links between themselves and others. Observational approaches struggle because of the non-stationarity of causal links in dynamic environments, and the sparsity of causal interactions while requiring the approaches to work in an online fashion. Meanwhile interventional approaches are impractical as a vehicle cannot experiment with its actions on a public road. To counter the issue of non-stationarity we reformulate the problem in terms of extracted events, while the previously mentioned restriction upon interventions can be overcome with the use of counterfactual simulation. We present three variants of the proposed counterfactual causal discovery method and evaluate these against state of the art observational temporal causal discovery methods across 3396 causal scenes extracted from a real world driving dataset. We find that the proposed method significantly outperforms the state of the art on the proposed task quantitatively and can offer additional insights by comparing the outcome of an alternate series of decisions in a way that observational and interventional approaches cannot.

\end{abstract}


\copyrightnotice


\section{INTRODUCTION} \label{sec:introduction}

In order for autonomous vehicles to operate in the wider world they must be capable of understanding the behavioural interactions between themselves and others. One can frame this problem as a process of discovering causal links between the decisions of agents. Causal Discovery (CD) approaches usually rely upon intervention based techniques resembling experimentation, or observational techniques which rely upon gathering sufficient data to provide insights on causal relationships. The former is impractical since it relies upon experimentation, which cannot be done in safety critical domains such as driving. The latter set of methods work well provided a sufficient quantity of data is present, the causal relationships are well represented within the data, and the causal relationships represented by the data are temporally stationary. Within an online driving setting comprised of multiple agents it is difficult to maintain these conditions, and our previous work has demonstrated the resulting performance degradation \cite{howard2023evaluating}.

In light of these limitations, we propose a counterfactual approach to CD with the aim of finding causal relationships between the behaviour of drivers. A counterfactual approach combines the discovery power of intervention based approaches without the need to actually experiment in the real world. This is achieved by carrying out the interventions and resulting analysis within simulation. The requirement to be able to simulate the agent's environment is a limitation of this type of approach, however implementing such a simulator in practice can rely upon a relatively small but well established base of knowledge.

\begin{figure}[t]
    \centering
    \includegraphics[width=0.7\linewidth]{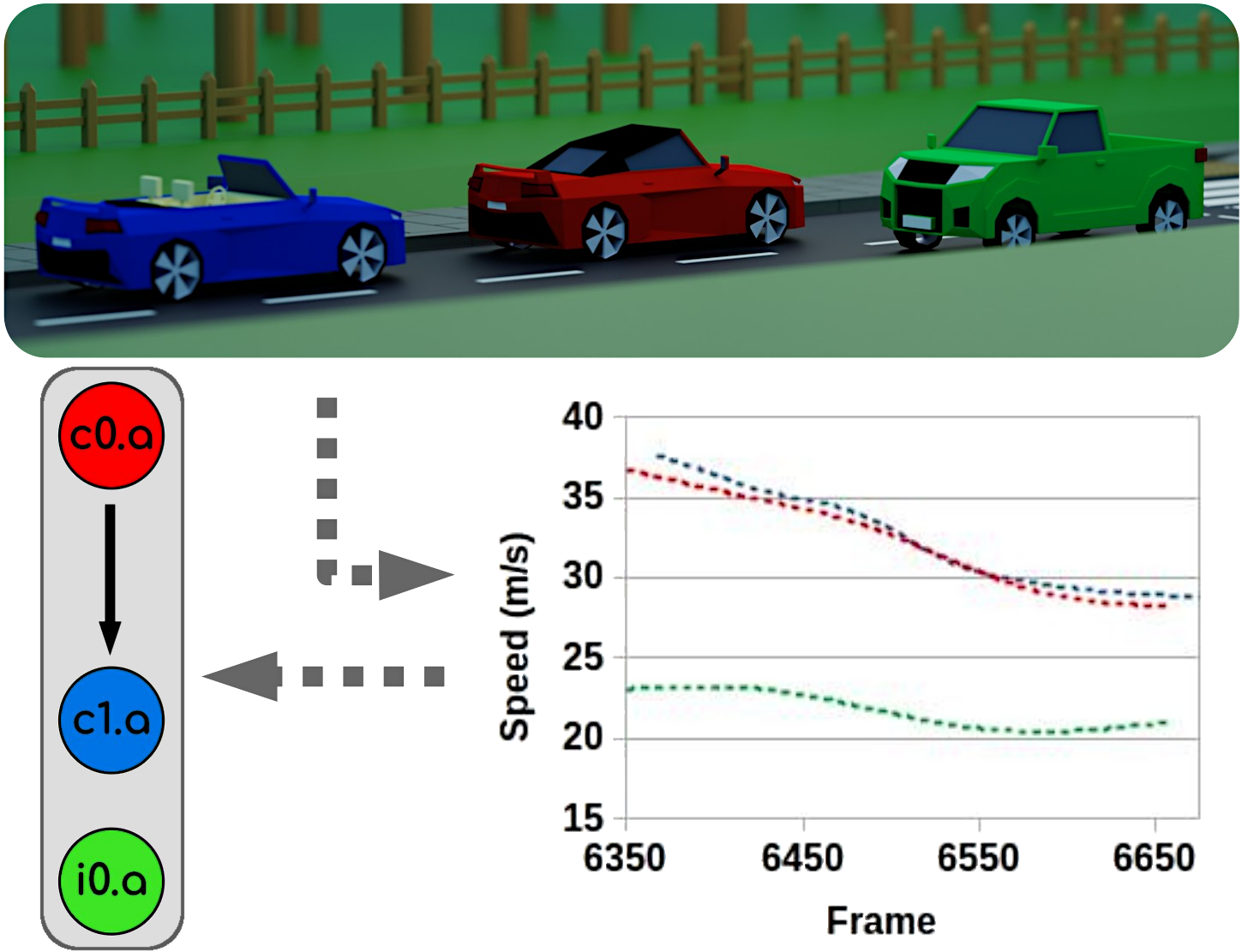}
    \caption{Illustrates a 3D depiction of the two vehicle convoy scenario considered in this paper, as well as a graph of example vehicle velocities over time for such a scenario, and the entity-level ground truth causal graph. Colour is consistent for each agent across elements.}
    \label{fig:driving_scenario}
\end{figure}

With the motivation and premise of our work established, we discuss the existing work in Sec. \ref{sec:related_work} before covering the following contributions with the rest of the paper:
\begin{itemize}
    \item A novel means of extracting agent target speed decisions.
    \item A novel CD method utilising counterfactual reasoning.
    \item Evaluation of proposed method against State of the Art.
\end{itemize}


\section{RELATED WORK} \label{sec:related_work}

\subsection{Causal Reasoning \& Discovery}

Causal reasoning as a formal area of study began to take hold through the work of Pearl \textit{et al.} \cite{pearl2000causality} based upon Pearl's earlier work upon Bayesian networks \cite{pearl1985bayesian} and expanding upon previously established ideas such as Granger causality \cite{granger1969investigating}. Later approaches were developed to discover causal models, with some of the first being the constraint-based PC algorithm developed by Spirtes \textit{et al.} \cite{spirtes2001causation} and the noise-based LiNGAM approach presented by Shimizu \textit{et al.} \cite{shimizu2006linear}. A comprehensive review of structural CD methods utilising observational data is provided by Glymour \textit{et al.} \cite{glymour2019review}. Lastly, while interventional approaches such as those proposed by Kocaoglu \textit{et al.} \cite{kocaoglu2017experimental} exist, they are not always practical depending upon the domain and task at hand.


\subsection{Temporal Causal Discovery}

When working within certain domains --- such as autonomous agents \slash vehicles --- accounting for temporal information associated with variables is necessary and \slash or desirable. Assaad \textit{et al.} \cite{assaad2022survey} provide a comprehensive review of observation-based temporal CD techniques and evaluate them upon synthetic data. In general these methods perform well upon auto-causation, but perform moderately or poorly upon cross-causation. Given that we are interested in the behavioural influence of agents upon one another, this latter category is critical to our work.
In order to evaluate the state of the art temporal CD methods within the Autonomous Driving (AD) domain our previous work consisted of carrying out a benchmark upon observational temporal CD methods \cite{howard2023evaluating}, utilising two real world driving datasets \cite{houston2020one,krajewski2018highd} and a synthetically generated dataset. We generally found that the approaches struggled due to the non-stationarity exhibited by agent causal relations and the sparsity with which causal interactions occurred, all the while working with a small time window of data.


\subsection{Counterfactual-Based Causal Discovery}
The primary reason for the scarcity of counterfactual CD is the difficulty of obtaining counterfactual data without first having a causal model to derive it from. Some approaches attempt to utilise a mix of observational and interventional data in order to approximate counterfactual data \cite{mueller2022causes} which is incorporated into the counterfactual-based CD process. 
However these methods still do not use counterfactual data, and are therefore reliant upon observational data given that interventions are impractical for the AD domain.
Hannart \textit{et al.} \cite{hannart2016causal} presents one of the few works that can produce counterfactual data as part of CD, and does so with a climate simulator in order to test for a causal link between historic greenhouse gas emmissions and the 2003 European heatwave. While simulators can make some assumptions regarding the ground truth climate model, they can ultimately be used to test for the presence of causal links between events and variables not inherently represented in the simulation framework. In this work we rely upon a kinematic simulator and an assumed theory of mind for each agent, yet no direct link exists between the decisions of one agent and another.

\subsection{Causality in Autonomous Agents \& Driving}
Several publications have identified causality as an important component in being able to explain and interpret AI \slash robotics \cite{gunning2019darpa, hellstrom2021relevance}. Recent work has theorised how causal reasoning might be integrated as part of Agent-Based Models (ABMs) \cite{istrate2021models}. Valogianni \& Padmanabhan \cite{valogianni2022causal} go as far as to implement an ABM that leverages elements of CD. Their work relies upon possessing a well defined set of agent rules that dictate behaviour and then using a genetic algorithm based technique to learn the parameterisation of these rules and their attribution towards an observed outcome.


One such research work is CausalCity \cite{mcduff2022causalcity}, which does not utilise causal reasoning but instead generates scenarios exhibiting realistic vehicle following behaviour for the purposes of benchmarking CD techniques. They evaluate the quality of the generated data by applying three graphical neural network based CD techniques. 
In contrast to CausalCity, CausalAF \cite{ding2022causalaf} aims to create safety critical driving scenarios for the general purpose of AD research but uses a causal model in the process.

In terms of AD research which aims to utilise CD, the works of de Haan \textit{et al.} \cite{dehaan2019causal} and Samsami \textit{et al.} \cite{samsami2021causal} both attempt imitation learning and simultaneously learn an embedded causal model in the process. However, the aim of utilising a causal model is to improve the imitation learning process rather than an informative model aimed to offer an insight into agent interactions within an AD scene, differentiating it from our work.

Finally, examples of CD being utilised to provide an interpretable causal model are given in the work of Kim \& Canny \cite{kim2017interpretable} and Li \textit{et al.} \cite{li2020who}. Both of these focus upon linking the behaviour of a vehicle to the data captured by its sensors, primarily in the form of camera images \slash video, as opposed to discovering causal links between the decisions of agents at an abstract level. 


\section{PROBLEM DEFINITION} \label{sec:problem_definition}
As discussed in Sec. \ref{sec:introduction} this work focuses upon examining causal links between driving agents that described behavioural interactions. Here we borrow the simple scenario that was used to examine this problem in our previous work \cite{howard2023evaluating}. This scenario describes a two car convoy with an additional independent vehicle as depicted in Fig. \ref{fig:driving_scenario}. As the lead vehicle in the convoy adjusts its velocity, we expect the tail vehicle to roughly mirror it, hence there should be a causal link discovered from the lead vehicle to the tail vehicle, meanwhile there should be no links to or from the independent vehicle. While this is quite a simplistic view of the causal interactions within the scenario the methods presented here are capable of much greater detail. For example, the work presented here can discover causal links between individual agent decisions as well as the type of behavioural reaction said links might resemble. However, it is precisely because these capabilities are absent from existing temporal CD techniques that we avoid this level of detail in quantitative evaluation in order to make this work comparable with the state of the art.


\section{BACKGROUND} \label{sec:background}

\subsection{Causal Graphs \& Discovery}
A causal graph $G = (V, L)$ is comprised of a tuple of variables $V$ and causal links $L$. Individual variables are indicated with a superscript (i.e. $V^i \in V$), meanwhile individual causal links are comprised of variable pairs as follows: 
\begin{equation}
L \subseteq \{\ (V^i, V^j)\ \|\ V^i, V^j \in V\ \}
\end{equation}
A causal link between $V^i$ and $V^j$ denoted as $L^{i, j}$, describes a relationship whereby a change in variable $V^i$ can affect variable $V^j$.

The typical goal of causal discovery is to produce and estimate of this graph $\hat{G} = (\hat{V}, \hat{L})$. Many methods make an assumption of causal sufficiency, or in other words assume that the variables provided at the start of causal discovery are sufficient to describe all causal activity present between those variables. In this situation the task becomes an attempt to estimate $\hat{L}$.

\subsection{Temporal Extension}
Across many situations, we might have values taken for a variable at multiple time steps. In this case it is useful to represent the variables as time series, $V^i = {[\ V^i_t\ ]}_{t \in T}$ where $T$ describes the time window variable values were captured within. Given that $V^i_t$ describes the $i$-th variable at a time $t$, we can also define the value of all variables at time $t$ as $V_t = {[\ V^i_t\ ]}_{0 \leq i \leq n}$.  Under this extension of the variables it is intuitive to redefine the overall variable set $V = {[\ V^i_t\ ]}_{0 \leq i \leq n,t \in T}$ in the process changing the relationship between $V$ and individual variables $V^i$ to $V^i \subseteq V$ rather than $V^i \in V$. Meanwhile we also can redefine the causal links to include time lags:
\begin{equation}
L \subseteq \{\ (V^i, V^j, t - t^\prime)\ \|\ V^i_t, V^j_{t^\prime} \in V, t \geq t^\prime\ \}
\end{equation}
where a given link $L^{i, j}_\tau$ describes a causal relationship whereby a change in variable $V^i$ affects variable $V^j$ following a time lag of $\tau$. Many methods make an assumption of causal stationarity, or in other words that $L$ does not itself change with time. The following section breaks from this somewhat, but it is nevertheless important as many of the methods we compare against do make this assumption.

\subsection{Shift to Event-Based Perspective}
Working with variables represented through a continuous time series is useful and appropriate for many fields where we wish to establish long term causal relationships utilising a large pool of data in an offline fashion (e.g. sociology, economics, medicine, etc.). However, counter to this is the circumstance where we wish to discover volatile causal relationships using limited data in an online fashion, which is often the case when working with autonomous agents that must interact with other agents. Not only do these situations make it difficult to assume causal stationarity, but they usually exhibit few cases of the causal relationships of interest being exhibited.

A better system in these circumstances is to think in terms of events occurring at discrete time points, and for testing for causal links between these events. Under this framework an event $E^i_t \in E \subseteq V \times T$ describes an event relating to the $i$-th variable occurring at time $t$. 
The events themselves just represent the assigning of a persisting value to the relevant variable at the specified time, however attempting to discover causal links between any and all cases where variables have been assigned a new value is intractable, which is why the problem must reformulated by assuming a theory of mind which agents operate under.

\subsection{Agent Theory of Mind} \label{subsec:agent_theory_of_mind}

\begin{figure}[t]
    \centering
    \includegraphics[width=0.8\linewidth]{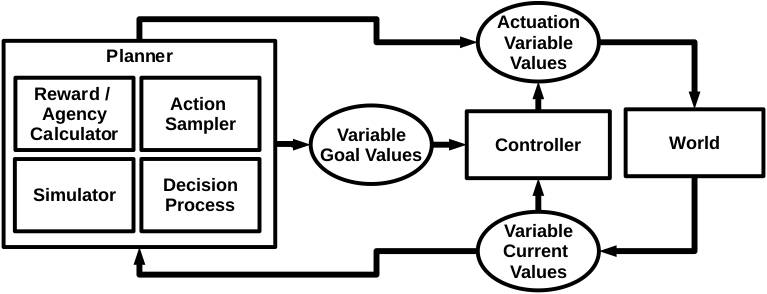}
    \caption{Illustration of the theory of mind ascribed to the way agents interact with the world. Circles represent data, whereas squares represent processes.}
    \label{fig:agent_world_interaction}
\end{figure}

In Fig. \ref{fig:agent_world_interaction} we illustrate the theory of mind we assume acts as a basis for an agent's behaviour. Namely the variables of the agent are divided into subsets $\mathcal{B}$, $\mathcal{G}$, and $\mathcal{A}$, denoting the base variables, goal variables, and actuation variables respectively. Base variables consist of variables that describe the state of the world as it is, and cannot be directly changed by an agent. Goal variables consist of variables that describe the state of the world the agent aims to achieve. The values of goal variables consist of both the goal target value for the base variable associated with the goal variable, as well as a goal target time.
For a given time $t$ this is denoted as $\mathcal{G}^i_t = ({\mathcal{B}^i_t}^\ast, t^{\prime}) \in \mathcal{B} \times T$. We make the assumption that the values of goal variables reflect reasoned decisions, either made by an AI planner or a human. Lastly, actuation variables represent values that can be specified by the agent, and are either assigned directly by the agent or via a controller utilising the actuation variables in order to achieve the goal state, however the former of these is not explored within this work, due to a lack of relevant data. 

To frame these variable cases in the context of the driving problem considered in this paper, we provide the following examples. The speed of a vehicle could be considered a base variable, as it describes the actual state of the world and cannot be directly changed by an agent. Meanwhile the desired speed of the vehicle would be a goal variable and would be selected by the agent as part of its planning \slash reasoning. Finally the acceleration of the vehicle would be an actuation variable as it is a variable which the agent can control in an effort to reach the desired speed. 
Finally, an example of a directly assigned actuation variable for a vehicle is whether the headlights or horn are active, as these can be toggled at whim.

By considering only goal variable events which correspond to reasoned decisions $D \subset E$ of the agent, we can redefine the target of our causal discover to be to find $\hat{G} = (\hat{D}, \hat{L})$. As such we are discovering a graph of causal links between agent decisions, rather than between variables, and therefore we can redefine $L$ as follows:
\begin{equation} \label{eq:decision_causal_links}
    L \subseteq \{\ ((\mathcal{G}^i_t, t), (\mathcal{G}^j_{t^\prime}, t^\prime)) \in D \times D\ \|\ t < t^\prime\ \}
\end{equation}
Given that each event is associated with a time, links between decisions $i$ and $j$ can be denoted as $L^{i,j}$ and henceforth this notation will be used for causal links. Since all of these events correspond to a variable, this representation can easily be converted to a variable-based or entity-based causal graph. For the purposes of this work, given two agents, if and only if there is a causal link from a decision of one agent to a decision of the other agent, there exists a causal link from the first agent to the second in the entity-based causal graph.

\section{AGENT DECISION EXTRACTION} \label{sec:goal_variable_event_extraction}
Before we can attempt to discover causal links $\hat{L}$ we must first extract the relevant decisions $\hat{D}$ from the variables represented as time series. While the counterfactual CD process is largely domain independent, this step does require a level of understanding of the domain. Here we consider the current and desired speed of an agent as the base and goal variables of interest, and the acceleration as the actuation variable of interest. However, the method presented here for extracting decisions can be applied to variables in any domains which possess a similar relationship. 

The first step to identifying goal variable events is to find a set of potential decision times $\mathcal{S}_{\mathcal{G}^{i}}$ and a set of potential goal target times $\mathcal{F}_{\mathcal{G}^{i}}$, which are derived as follows:
\begin{equation}
\begin{split}
    \mathcal{S}_{\mathcal{G}^{i}} = \{  &\  t \in T\ \|\ (\mathcal{A}^{\mathcal{B}^i}_t < \lambda_{\mathcal{A}^{\mathcal{B}^i}} \land \mathcal{A}^{\mathcal{B}^i}_{t+\Delta_t} \geq \lambda_{\mathcal{A}^{\mathcal{B}^i}})\ \lor \\
                        & \quad(\mathcal{A}^{\mathcal{B}^i}_t > -\lambda_{\mathcal{A}^{\mathcal{B}^i}} \land \mathcal{A}^{\mathcal{B}^i}_{t+\Delta_t} \leq -\lambda_{\mathcal{A}^{\mathcal{B}^i}}), \\
                        & \quad\mathcal{A}^{\mathcal{B}^i}_t, \mathcal{A}^{\mathcal{B}^i}_{t+\Delta_t} \in \mathcal{A}\ \}\ \cup\ \min\ T
\end{split}
\end{equation}
\begin{equation}
\begin{split}
    \mathcal{F}_{\mathcal{G}^{i}} = \{  &\  t \in T\ \|\ (\mathcal{A}^{\mathcal{B}^i}_t \geq \lambda_{\mathcal{A}^{\mathcal{B}^i}} \land \mathcal{A}^{\mathcal{B}^i}_{t+\Delta_t} < \lambda_{\mathcal{A}^{\mathcal{B}^i}})\ \lor \\
                        & \quad(\mathcal{A}^{\mathcal{B}^i}_t \leq -\lambda_{\mathcal{A}^{\mathcal{B}^i}} \land \mathcal{A}^{\mathcal{B}^i}_{t+\Delta_t} > -\lambda_{\mathcal{A}^{\mathcal{B}^i}}), \\
                        & \quad\mathcal{A}^{\mathcal{B}^i}_t, \mathcal{A}^{\mathcal{B}^i}_{t+\Delta_t} \in \mathcal{A}\ \}\ \cup\ \max\ T
\end{split}
\end{equation}
where $\lambda_{\mathcal{A}^{\mathcal{B}^i}}$ is a threshold used to determine there has been sufficient actuation in order to conclude a decision could have been made.
Assuming that $\mathcal{S}_{\mathcal{G}^{i}}$ and $\mathcal{F}_{\mathcal{G}^{i}}$ are sorted and can be indexed, Algorithm \ref{alg:agent_decision_extraction} describes the process by which the set of agent decisions on the goal variable $\hat{D}^{\mathcal{G}^{i}}$ can be extracted. Here $\lambda_{\delta t}$ is a threshold used to ensure there is enough time between the decision time and the goal target time in order to reflect an effort from the agent over a protracted length of time. Meanwhile $\lambda_{\delta \mathcal{B}^i}$ is used to determine whether there is a sufficient difference between the base variable value at decision time and the goal target value in order to be significant. Assuming both of these conditions are met, an additional decision is added to $\hat{D}^{\mathcal{G}^{i}}$, otherwise the next potential goal target time is considered. Finally, whenever a decision is added to $\hat{D}^{\mathcal{G}^{i}}$, the next potential decision time is selected from those greater than or equal to the goal target time of the recently added decision, as only one goal can be pursued at a time for any given base variable.

\begin{algorithm}[t] 
\caption{Agent Decision Extraction} \label{alg:agent_decision_extraction}
\footnotesize
\KwData{$T$, $\mathcal{B}^i \in \mathcal{B} \subset V$,\ $\mathcal{S}_{\mathcal{G}^{i}} \subseteq T$,\ $\mathcal{F}_{\mathcal{G}^{i}} \subseteq T$}
\KwResult{$\hat{D}^{\mathcal{G}^{i}} \subset (V \times T) \times T$}
$\hat{D}^{\mathcal{G}^{i}} = \{\}$,\ $j = 0$,\ $k = 0$\;
\While{$j < |\mathcal{S}_{\mathcal{G}^{\mathcal{B}^i}}| \land k < |\mathcal{F}_{\mathcal{G}^{\mathcal{B}^i}}|$}{
    $t = \mathcal{S}_{\mathcal{G}^{i}}[j]$,\ $t^\prime = \mathcal{F}_{\mathcal{G}^{i}}[k]$\;
    \eIf{$t^\prime - t \geq \lambda_{\delta t} \land |\mathcal{B}^i_t - \mathcal{B}^i_{t^\prime}| \geq \lambda_{\delta \mathcal{B}^i}$}{
        $\hat{D}^{\mathcal{G}^{i}}_t = ((\mathcal{B}^i_{t^\prime}, t^\prime), t)$\;
        $\hat{D}^{\mathcal{G}^{i}} = \hat{D}^{\mathcal{G}^{i}} \cap \{ \hat{D}^{\mathcal{G}^{i}}_t \}$\;
        \While{$\mathcal{S}_{\mathcal{G}^{i}}[j] < \mathcal{F}_{\mathcal{G}^{i}}[k]$}{
            $j$++\;
        }
    }{
        $k$++\;
    }
}
\If{$|\hat{D}^{\mathcal{G}^{i}}| == 0$}{
    $t = \min\ T$\;
    $\hat{D}^{\mathcal{G}^{i}}_t = ((\mathcal{B}^i_{t}, t), t)$\;
    $\hat{D}^{\mathcal{G}^{i}} = \hat{D}^{\mathcal{G}^{i}} \cap \{ \hat{D}^{\mathcal{G}^{i}}_t \}$\;
}
\end{algorithm}


\section{COUNTERFACTUAL TEMPORAL CAUSAL DISCOVERY} \label{sec:counterfactual_temporal_causal_discovery}
\subsection{Potential Causal Link Iteration}
Provided we can provide a means for testing for the presence of individual causal links, we can simplify the problem of causal discovery to iterating over potential causal links and applying the aforementioned test. With this in mind, we can define the potential causal links as follows:
\begin{equation}
    \tilde{L} = \{\ ((\mathcal{G}^i_t, t), (\mathcal{G}^j_{t^\prime}, t^\prime)) \in D \times D\ \|\ i \neq j,\ t < t^\prime\ \}
\end{equation}
This definition differs slightly from \ref{eq:decision_causal_links} in that it does not include links between the decisions of the same agent. This is because we are primarily interested in how the behaviour of agents affects the decision making process of other agents and it can safely be assumed that an agent's past actions will influence its future actions. Furthermore, for the purposes of clarity, the first element $(\mathcal{G}^i_t, t)$ (i.e. the cause) will henceforth be denoted $\mathcal{C}$ while the second element $(\mathcal{G}^j_{t^\prime}, t^\prime)$ will now be denoted $\mathcal{E}$.

\subsection{Counterfactual Simulation}
Before we can evaluate whether or not a causal link is likely present we first require a means by which to acquire counterfactual data. As discussed in Sec. \ref{sec:related_work} typically we require a causal model in order to acquire counterfactual data. This is true of the simulation we utilise as well, however the causal links we wish to discover are the abstract behavioural causal links between agent decisions. Given that we assume the world follows basic kinematics and the agents follow the previously described theory of mind, we can construct a simulation that embeds a low-level causal model, from which we can gather counterfactual data to use in the process of high-level causal discovery (i.e. between agent decisions).

Given the theory of mind depicted in Fig. \ref{fig:agent_world_interaction}, the three processes that are required to simulate the variables associated with an agent are its planner, controller and a means of simulating the interactions of the agents with their environment through their actuation and base variables.

The planner represents the decision-making process of an agent and therefore constitutes the component we are interested in for the purposes of causal discovery. This is detailed in Sec. \ref{subsec:counterfactual_causal_link_testing}, while the controller and world simulation components are documented below.

\subsubsection{Controller}
For a given agent the vehicular acceleration actuation variable $\mathcal{A}^a$ is assumed to be calculated directly from a proportional error based calculation. Given that the vehicular velocity at time $t$ is given as $\mathcal{B}^v_t$ and the goal is given as $\mathcal{G}^v_t = ({\mathcal{B}^v_t}^\ast, t^\prime)$, we can calculate $\mathcal{A}^a_t$ as follows:
\begin{equation}
    \mathcal{A}^a_t = \frac{(\mathcal{B}^v_t - {\mathcal{B}^v_t}^\ast)}{\max\ \{ t^\prime - t, \Delta_t \}}
\end{equation}
In other words the acceleration is set in order to achieve the change in velocity required by the goal velocity target within the time frame dictated by the goal time target. Should the goal time target have been passed the controller will set the acceleration in order to try and achieve the goal velocity target within a single time step. While proportional error based approaches can be prone to oscillation and more complex models can be used to overcome this, we found that in our own experimentation, the calculation given above was sufficient to produce satisfactory results.

\subsubsection{World Simulation}
One of the benefits of working within the AD domain is that the world and the interactions of agents with it can largely be explained through kinematics \slash dynamics. As a result we represent the world through a series of 2D kinematic equations that treat vehicles as rigid body particles with rotation \slash angular velocity. It should be noted that while these angular components are present in the simulation, they play a small role due to the documented experiments taking place upon straight stretches of road. The simulation assumes a fixed time step $\Delta_t$ when making calculations. Additionally the simulator makes use of a collision detection system utilising radius, bounding box and Separating Axis Theorem based checks.

\subsection{Counterfactual Causal Link Testing} \label{subsec:counterfactual_causal_link_testing}
Once again referencing Fig. \ref{fig:agent_world_interaction}, we can see that the planner takes its input from the base variables of an agent and outputs to the actuation and goal variables of an agent. As stated in Sec. \ref{subsec:agent_theory_of_mind} we do not consider direct assignment of actuation variables by the planner in this work due to the lack of available data demonstrating this phenomena. Therefore for the purposes of this work, the causal effect of the planner on actuation variables directly is assumed to be non-existent, and the link is thus removed.

In the process of examining the causal links between decisions we need to carry out interventions upon the decisions made by the agents of interest for a given causal link and observe the difference in outcome. Most importantly we want to examine the planners perception of the outcome through the reward \& agency calculators documented below. The concept is that since we assume the affected agent wishes to maximise its reward and maintain its agency. Generally we expect that the affected decision $\mathcal{E}$ is more advantageous in the circumstance the causing decision $\mathcal{C}$ has occurred. Furthermore we would expect that the affected decision $\mathcal{E}$ would not have been advantageous had causing decision $\mathcal{C}$ not occurred, otherwise the agent would likely take decision $\mathcal{E}$ regardless of whether $\mathcal{C}$ occurred. For the purposes of this paper we focus primarily upon necessary causal relations, or in other words causal relations where $\mathcal{E}$ would not have occurred if not for $\mathcal{C}$, rather than cases where $\mathcal{C}$ was sufficient to cause $\mathcal{E}$ but not necessary.

To derive counterfactual data regarding the reward \& agency of the affected agent for an intervened set of decisions, it is assumed that given a set of decisions $D$, an initial set of base variables $\mathcal{B}_{t_\alpha}$, a controller and a world simulator one can simulate an alternate series of variable values $\hat{V}_{\geq t_{\mathcal{C}}}$. This process is assumed to mirror the agent's own planning process and similarly we can run reward and agency calculations to determine which decisions are advantageous or disadvantageous. From these calculations we can then determine if the decision pair matches the aforementioned criteria for a causal link.

The four sets of decisions which we consider here are the original set, the set without the affected decision $\mathcal{E}$, the set without the causing decision $\mathcal{C}$ and the set without either $\mathcal{E}$ or $\mathcal{C}$. The values and sets derived from each simulation are denoted by the subscripted parenthesis ${(\cdot)}_{\mathcal{E},\mathcal{C}}$, ${(\cdot)}_{\neg\mathcal{E},\mathcal{C}}$, ${(\cdot)}_{\mathcal{E},\neg\mathcal{C}}$, and ${(\cdot)}_{\neg\mathcal{E},\neg\mathcal{C}}$ respectively. From here we present three approaches to evaluating the advantageous vs disadvantageous nature of the outcomes associated with each, a reward-based approach, an agency-based approach and a hybrid approach.

\subsubsection{Reward-Based Approach}
The reward-based approach makes the assumption there is some metric to measure how ``good" a particular world state is from the perspective of an agent. For the affected decision agent the reward at time $t$ is denoted $R_t^{\mathcal{E}} \in [0, 1]$, and is calculated as follows:
\begin{equation}
    R_t^{\mathcal{E}} = r_{ttc}(V_t^{ttc}) \times r_{cct}(V_t^{cct}) \times r_v(V_t^v)
\end{equation}
\begin{equation}
    r_{ttc}(ttc) = 1 - e^{-ttc}
\end{equation}
\begin{equation}
    r_{cct}(cct) = 
    \begin{cases}
        1, & cct > 0\\
        0, & cct \leq 0
    \end{cases}
\end{equation}
\begin{equation}
    r_v(v) = 1 - 0.5 \times e^{-\max \{0.1 \times v,\,0\}}
\end{equation}
where $ttc$, $cct$ and $v$ refer to the Time To Collision (TTC), Cumilative Collision Time (CCT) and speed of the affected decision agent respectively. Using this metric we can calculate the difference between the minimum reward after the affected decision time for the original set of decisions and the set of decisions without the affected decision:
\begin{equation}
    \delta R^\mathcal{E}_{+} = \min_{t \in (t_{\mathcal{E}}, t_\Omega]} {(R_t^{\mathcal{E}})}_{\mathcal{E},\mathcal{C}} - \min_{t^\prime \in (t_{\mathcal{E}}, t_\Omega]} {(R_{t^\prime}^{\mathcal{E}})}_{\neg\mathcal{E},\mathcal{C}}
\end{equation}
where $t_{\Omega} = \min \{ \max T^\mathcal{E} , \max T \}$ and $T^\mathcal{E}$ is the time window the agent associated with $\mathcal{E}$ was captured for. This metric is indicative of the perceived advantage of taking decision $\mathcal{E}$ following $\mathcal{C}$. In contrast to this we can also calculate the difference between the minimum reward after the affected decision time for the set of decisions without the causing \& affected decisions, and the set of decisions just without the causing decision:
\begin{equation}
    \delta R^\mathcal{E}_{-} = \min_{t \in (t_{\mathcal{E}}, t_\Omega]} {(R_t^{\mathcal{E}})}_{\neg\mathcal{E},\neg\mathcal{C}} - \min_{t^\prime \in (t_{\mathcal{E}}, t_\Omega]} {(R_{t^\prime}^{\mathcal{E}})}_{\mathcal{E},\neg\mathcal{C}}
\end{equation}
This metric on the other hand is indicative of the perceived disadvantage of taking decision $\mathcal{E}$ in the absence of $\mathcal{C}$. The combination of these is given by $\delta R^\mathcal{E} = \delta R^\mathcal{E}_{+} + \delta R^\mathcal{E}_{-}$. From here we can threshold the metric to determine whether or not a causal link is present:
\begin{equation}
    \xi_{R} (L^{\mathcal{C}, \mathcal{E}}) = 
    \begin{cases}
        1, & \delta R^\mathcal{E} \geq \lambda_{\delta R}\\
        0, & \delta R^\mathcal{E} < \lambda_{\delta R}
    \end{cases}
\end{equation}
where $\lambda_{\delta R}$ is a predefined threshold on the reward-based decision dependent advantage \slash disadvantage metric.

\subsubsection{Agency-Based Approach}
Agency in the context of this work refers to an agent's ability to carry out its decisions effectively. For a given time $t$ the agency of the agent associated with the affected decision is denoted by $A_t^{\mathcal{E}} \in \{0, 1\}$. The most common example of this property being violated is as a result of a collision between agents, although environmental factors such as wind and ice could also influence this however these are harder to capture. With this in mind we define $A_t^{\mathcal{E}}$ as follows:
\begin{equation}
    A^{\mathcal{E}} = \bigvee_{t \in (t_{\mathcal{E}}, t_\Omega]} 
    \begin{cases}
        1, & V_{t}^{{cct}_{\mathcal{E}}} > 0 \land V_{t}^{{cct}_{\mathcal{C}}} > 0\ \land\\
        & V_{t-\Delta_t}^{{cct}_{\mathcal{E}}} \leq 0 \land V_{t-\Delta_t}^{{cct}_{\mathcal{C}}} \leq 0\\
        0, & \text{otherwise}
    \end{cases}
\end{equation}
where $V_{t}^{{cct}_{\mathcal{E}}}$ and $V_{t}^{{cct}_{\mathcal{C}}}$ refer to the cumulative collision times for the affected decision agent and causing decision agent respectively. It is assumed that an agent will act in such a way as to maintain their ability to enact their own decisions. Therefore now we have a way of calculating agency, we can describe four agency patterns which pertain to causal relationships or the lack thereof:
\begin{equation}
    \xi_{A}^{\text{active}} (L^{\mathcal{C}, \mathcal{E}}) = \neg{(A^{\mathcal{E}})}_{\mathcal{E},\mathcal{C}} \land {(A^{\mathcal{E}})}_{\neg\mathcal{E},\mathcal{C}} \land \neg{(A^{\mathcal{E}})}_{\neg\mathcal{E},\neg\mathcal{C}}
\end{equation}
\begin{equation}
    \xi_{A}^{\text{passi.}} (L^{\mathcal{C}, \mathcal{E}}) = \neg{(A^{\mathcal{E}})}_{\mathcal{E},\mathcal{C}} \land {(A^{\mathcal{E}})}_{\mathcal{E},\neg\mathcal{C}} \land \neg{(A^{\mathcal{E}})}_{\neg\mathcal{E},\neg\mathcal{C}}
\end{equation}
\begin{equation}
    \xi_{A}^{\text{facil.}} (L^{\mathcal{C}, \mathcal{E}}) = \neg{(A^{\mathcal{E}})}_{\mathcal{E},\mathcal{C}} \land {(A^{\mathcal{E}})}_{\mathcal{E},\neg\mathcal{C}} \land {(A^{\mathcal{E}})}_{\neg\mathcal{E},\neg\mathcal{C}}
\end{equation}
\begin{equation}
    \xi_{A}^{\text{m.e.m.}} (L^{\mathcal{C}, \mathcal{E}}) = \neg{(A^{\mathcal{E}})}_{\mathcal{E},\mathcal{C}} \land {(A^{\mathcal{E}})}_{\neg\mathcal{E},\mathcal{C}} \land {(A^{\mathcal{E}})}_{\neg\mathcal{E},\neg\mathcal{C}}
\end{equation}
The active and passive cases describe situations where a causal link should be present. The active case describes a situation where the decision $\mathcal{E}$ must be taken given that $\mathcal{C}$ has occurred in order to avoid a loss of agency, or in other words the agent is actively forced to take the affected decision. The passive case in contrast describes a situation where taking decision $\mathcal{E}$ in the circumstance where $\mathcal{C}$ has not occurred would lead to a loss of agency, and thus the occurrence of $\mathcal{C}$ passively allows the decision $\mathcal{E}$ to be made. In contrast the facilitation and mutual effect motive cases describe cases where a causal link is unlikely to be present or impossible to test for. In the facilitation case, because agency is lost regardless of whether $\mathcal{E}$ is taken assuming $\mathcal{C}$ did not occur, it is impossible to evaluate the impact the cause's presence has upon the choice to take decision $\mathcal{E}$. Meanwhile the mutual effect motive describes a case where failing to take $\mathcal{E}$ regardless of whether $\mathcal{C}$ has occurred will lead to a loss of agency, and therefore it is in the agent's interest to take decision $\mathcal{E}$ in either case.

With these metrics defined, we can now determine whether a causal link is present:
\begin{equation}
    \xi_{A} (L^{\mathcal{C}, \mathcal{E}}) = (\xi_{A}^{\text{active}} \lor \xi_{A}^{\text{passi.}}) \land \neg (\xi_{A}^{\text{facil.}} \lor \xi_{A}^{\text{m.e.m.}})
\end{equation}

\subsubsection{Hybrid Approach}
Here we describe an approach that combines elements from the reward-based and agency-based approaches. Given the previously defined causal link tests and metrics, we can defined the hybrid test as follows:
\begin{equation}
    \xi_{H} (L^{\mathcal{C}, \mathcal{E}}) = (\xi_{A}^{\text{active}} \lor \xi_{A}^{\text{passi.}} \lor \xi_{R})\,\land 
    \neg (\xi_{A}^{\text{facil.}} \lor \xi_{A}^{\text{m.e.m.}})
\end{equation}
This corresponds to the agency-based causal link test with the additional opportunity to add a causal link if the reward-based metric is above the predefined threshold and not invalidated by an agency-based metric.


\section{EXPERIMENTS} \label{sec:experiments}

\subsection{Code \& Data}
The code associated with this work along with experiment data produced is available on GitHub\footnote{https://github.com/cognitive-robots/counterfactual-cd-paper-resources}. The input data is derived from the High-D dataset \cite{krajewski2018highd} which contains labelled agents recorded from stretches of highway in Germany. Scenes which exhibit causal behaviour are extracted from the dataset by finding pairs of vehicles in convoy that exhibit a sufficient relative change in velocity. Meanwhile the independent vehicle is selected from a separate lane to the convoy in order to ensure its independence. This leads to a total of $3396$ causal scenes being extracted from the dataset.

\subsection{Evaluation Metrics}
The work presented here is evaluated for two metrics, the CD efficacy and the computational performance measured in time elapsed during causal discovery. The CD efficacy is measured in terms of $\text{F}_1$ Score, Precision and Recall. These values are calculated in the same manner to our previous work \cite{howard2023evaluating} whereby the links or lack thereof contribute towards true and false positives, or true and false negatives respectively.

\subsection{Parameters}
The three thresholds used by the decision extraction process are fixed with values of $0.2\, m/s^2$, $1.0\,s$, and $1\,m/s$ used for $\lambda_{\mathcal{A}^{\mathcal{B}^i}}$, $\lambda_{\delta t}$ and $\lambda_{\delta \mathcal{B}^i}$ respectively. These values were found to have satisfactory performance during preliminary experimentation, however formal evaluation of this is a goal for future work. The remaining parameter is the reward-based metric threshold denoted by $\lambda_{\delta R}$, for each batch of experiments this value is varied between $0.0$ and $1.0$ using an interval of $0.1$, with $0.01$ used in lieu of $0.0$.

\subsection{Baselines}
We compare the performance of the proposed method with the benchmarked performance of the state of the art on the same task \cite{howard2023evaluating}. For the purpose of clarity we present only the best three performing methods. These are DYNOTEARS \cite{pamfil2020dynotears} (speed-based, max. lag = $25\,s$, sig. = $0.05$), MVGC (acceleration-based, max. lag = $25\,s$, sig. = $0.1$) \cite{geweke1982measurement} and TiMINO \cite{peters2013causal} (acceleration-based, max. lag = $25\,s$, sig. = $0.1$). We also give the performance metrics given by assigning entity causal links with a probability of $0.5$ (i.e. random assignment).

\subsection{Results \& Discussion}

\begin{figure}[t]
    \centering
    \includegraphics[width=0.965\linewidth]{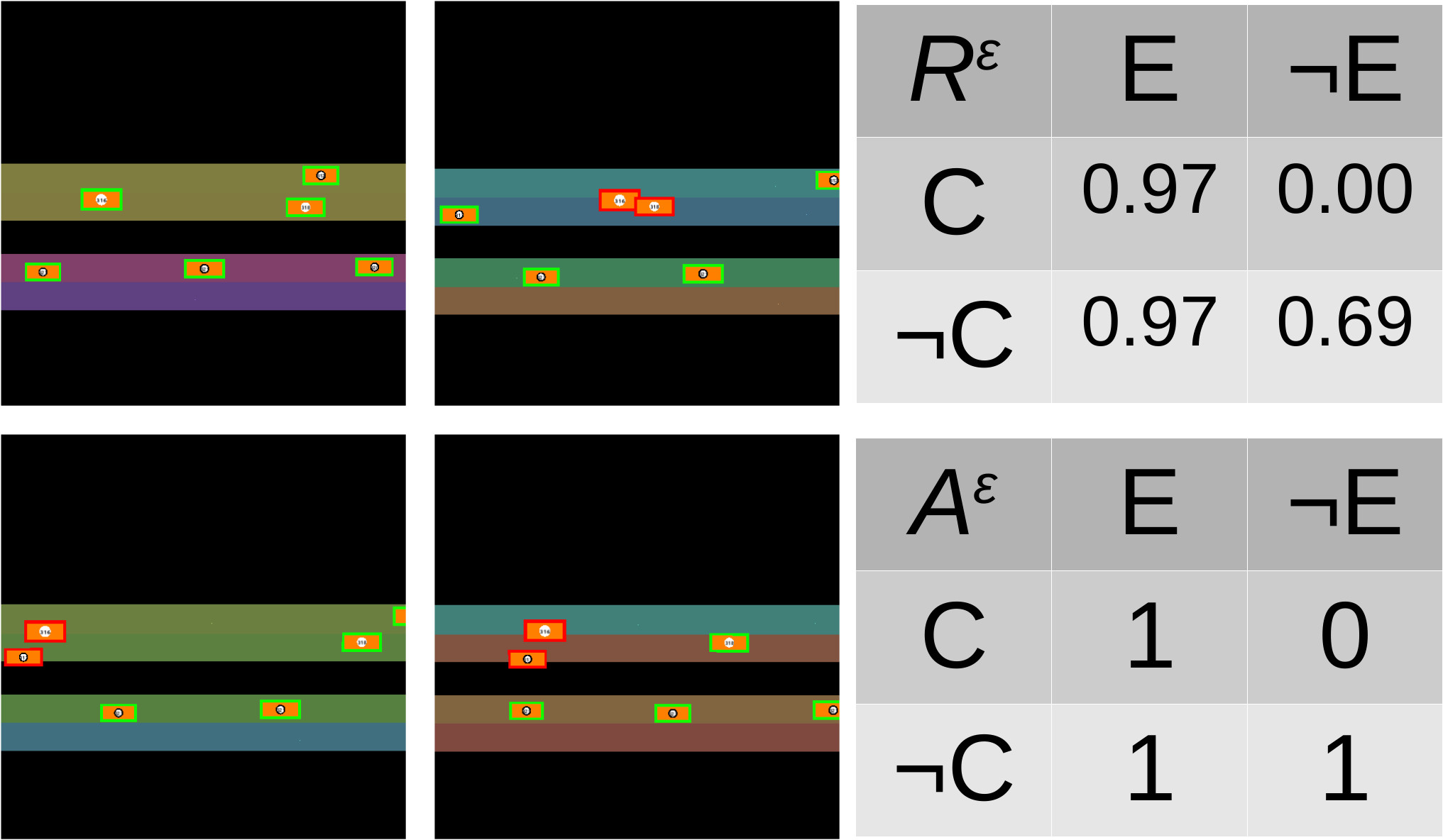}
    \caption{Illustrates the outcomes associated with the four sets of decisions considered for a single causal link test from the head convoy vehicle to the tail convoy vehicle. The convoy vehicles are indicated in the visualisation by the lack of black rims on the central circles of the agents. Note that the convoy tail only experiences a collision if $\mathcal{C}$ occurs and $\mathcal{E}$ does not.}
    \label{fig:qualitative}
\end{figure}

\begin{figure*}[t]
    \centering
    \includegraphics[width=0.87\linewidth]{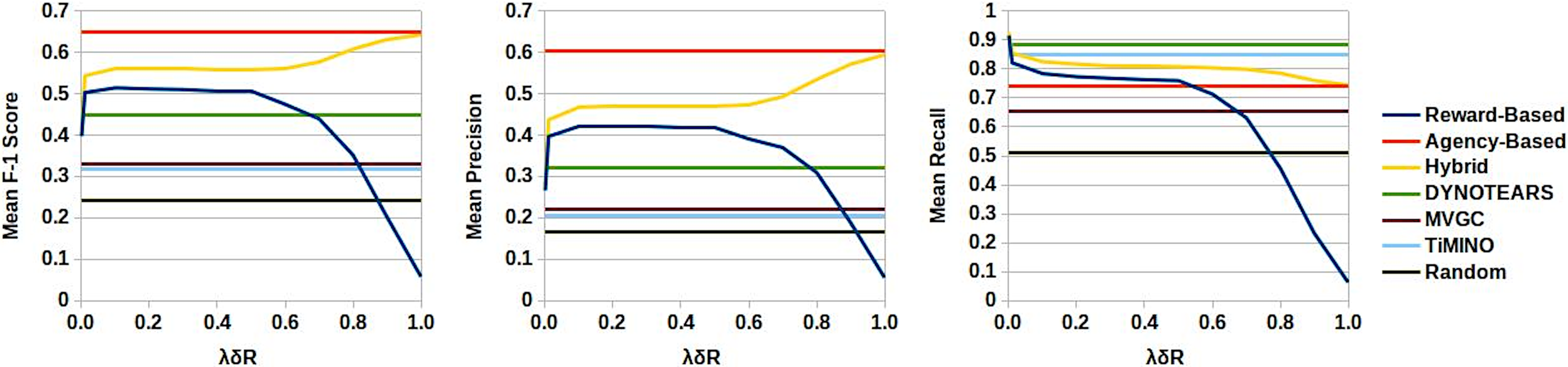}
    \caption{Results of applying the three variants of the counterfactual method and baselines to the 3396 causal scenes extracted from the High-D dataset.}
    \label{fig:results}
\end{figure*}

Results regarding the efficacy of the proposed CD approach variants is presented in Fig. \ref{fig:results}, while an illustration of the outcome comparison process is shown in Fig. \ref{fig:qualitative}. In general all of the variants outperform MVGC and TiMINO by a notable margin, with the hybrid and agency-based variants outperforming all three baselines. The peak mean $\text{F}_1$ score values for the reward-based, agency-based, and hybrid approaches are $0.514$, $0.649$, and $0.643$ respectively. The latter of these two outperform the best baseline (i.e. DYNOTEARS) by $0.2$, which is a notable margin considering there is a similar difference in $\text{F}_1$ score between DYNOTEARS and the entirely random selection of causal links.

The differences in performance between the reward-based and agency-based variants indicate that the simplistic collision driven nature of the agency calculations provide a strong starting point for this type of causal discovery. It was expected that the reward calculations based upon several factors would provide a more nuanced test which could capture edge cases that the agency-based variant could not. 
As such the hybrid variant was intended to combine the strengths of these two variants to compliment one another, however the evaluation metrics show that the agency-based variant significantly outperforms the reward-based variant.
The result is that the hybrid variant resembles a version of the agency-based variant which is only hindered by the inclusion of elements relating to the reward-based variant.
This is reflected in the fact that the hybrid variant's peak $\text{F}_1$ score is found when $\lambda_{\delta R} = 1.0$, or in other words when the barrier for a causal link to be accepted by the reward-based variant is at its highest. This is also explains why the hybrid variant increases in performance as the reward-based variant decreases in performance. While this does mean that for the purposes of this work the agency-based variant provides the best performance, the concept of a reward-based variant still holds merit in that it can better capture a variety of agent motivations across a continuous set of values. However, with this greater complexity comes a greater need for fine tuning, thus further research is required.

Finally, in terms of the execution times associated with each method, the counterfactual CD approach had a mean execution time of $3.16\,s$, compared with $0.407\,s$, $6.63\,s$, and $3.68\,s$ for the DYNOTEARS, MVGC, and TiMINO methods respectively. In other words, while DYNOTEARS can carry out causal discovery faster, the proposed method is still competitive when compared to MVGC and TiMINO. Furthermore $3\,\text{--}\,4\,s$ is not an unreasonable time for an agent to spend determining a causal relationships for a $10\,\text{--}\,30\,s$ scene, considering that if this method were to be used in an online fashion processing new data could be processed fast enough to update a causal model in real time.


\section{FUTURE WORK \& CONCLUSION} \label{sec:discussion}

A previously touched upon direction for future work would involve independently evaluating the performance of the counterfactual CD approach with ground truth decisions provided and the performance of the decision extraction process to correctly identify the ground truth decisions. This would help to disentangle the performance associated with each step of the overall methodology presented here.
A second direction for future work would be the continued development of the reward-based variant with the hope of bringing further performance improvements. A potential approach to this could be learning a reward model rather than relying upon assumptions of how a driver's decisions are motivated.
Beyond this, another direction would be to evaluate the method quantitatively across a number of different causal scenarios (e.g. three agent convoys, overtaking, merging) and examine the causal graphs discovered on these more complex scenarios qualitatively. In a similar fashion, further examination of CD failures on a link by link basis should facilitate further improvements in the method's performance. 
Lastly, while this work provides a means of discovering a causal model of road agent decisions, it does not specify how to apply the knowledge gained from this in future agent decisions. This limits the method's current use to retrospective scene analysis (e.g. road accident analysis).


In summary, traditional observation or intervention based CD approaches struggle when applied to real world driving scenarios. We have established that utilising simulation to derive counterfactual data is an effective means to carry out causal discovery between agent decisions. 
We argue that this work sets a strong precedence for the use of counterfactual CD for similar scenarios, both for its demonstrated performance, and for its ability to offer information on causal links derived from its ability to envision alternate outcomes.






\fontsize{8pt}{9pt}\selectfont
\bibliographystyle{IEEEtran}
\bibliography{references.bib}

\end{document}